\documentclass[letterpaper, 10 pt, conference]{ieeeconf}  

\usepackage{graphicx}
\usepackage{multirow}
\usepackage{algorithm}
\usepackage{subcaption}
\usepackage{soul}
\usepackage{comment}
\usepackage[noend]{algpseudocode}
\usepackage{afterpage}
\usepackage{array}
\usepackage{amsmath}
\usepackage{amsmath} 
\usepackage{balance}
\usepackage[font=small]{caption}
\usepackage{hyperref}
\usepackage{titlesec}

\IEEEoverridecommandlockouts                              

\title{\LARGE \bf
ImpedanceGPT: VLM-driven Impedance Control of Swarm of Mini-drones for Intelligent Navigation in Dynamic Environment
}

\author{Faryal Batool$^{*}$, Yasheerah Yaqoot$^{*}$, Malaika Zafar$^{*}$, Roohan Ahmed Khan, \\ Muhammad Haris Khan, Aleksey Fedoseev, and Dzmitry Tsetserukou %
\thanks{The authors are with the Intelligent Space Robotics Laboratory, Skolkovo Institute of Science and Technology. 
{\tt \{faryal.batool, yasheerah.yaqoot, malaika.zafar, roohan.khan, haris.khan, aleksey.fedoseev, d.tsetserukou\}@skoltech.ru}}
\thanks{*These authors contributed equally to this work.}
}

\begin{document}

\maketitle
\thispagestyle{empty}
\pagestyle{empty}
\vspace{-15pt} 
\begin{abstract}
Swarm robotics plays a crucial role in enabling autonomous operations in dynamic and unpredictable environments. However, a major challenge remains ensuring safe and efficient navigation in environments shared by both dynamic alive (e.g., humans) and dynamic inanimate (e.g., non-living objects) obstacles. In this paper, we propose \textbf{ImpedanceGPT}, a novel system that leverages a Vision-Language Model (VLM) with Retrieval-Augmented Generation (RAG) framework to enable real-time reasoning for adaptive navigation of mini-drone swarm in complex environments.
The key innovation of \textbf{ImpedanceGPT} lies in the merging VLM-RAG system with impedance control method, which is an active compliance strategy. This system provides drones with an enhanced semantic understanding of their surroundings and allows them to dynamically adjust impedance control parameters in response to obstacle types and environmental conditions. Our approach not only ensures safe and precise navigation but also improves coordination between drones in the swarm.
Experimental evaluations demonstrate the effectiveness of our system. The VLM-RAG framework achieved an obstacle detection and retrieval accuracy of \textbf{80\%} under optimal lighting. In static environments, drones navigated dynamic inanimate obstacles at \textbf{1.4 m/s} but slowed to \textbf{0.7 m/s} with increased safety margin around humans. In dynamic environments, speed adjusted to \textbf{1.0 m/s} near hard obstacles, while reducing to \textbf{0.6 m/s} with higher deflection region to safely avoid moving humans.



Video of ImpedanceGPT: \href{https://youtu.be/JTdeg9bAzL4}{https://youtu.be/JTdeg9bAzL4}
Github: \href{https://github.com/Faryal-Batool/ImpedanceGPT}{https://github.com/Faryal-Batool/ImpedanceGPT}
\end{abstract}
\textbf{\textit{Keywords:}} \textbf{\textit{ Vision-Language-Model, Impedance Control, Path Planning, Retrieval-Augmented Generation}}

\begin{figure}[t!]
\centering
\includegraphics[width=1\linewidth]{figures/main_image2.jpg}
\caption{\textbf{ImpedanceGPT} framework for adaptive swarm navigation. The system adaptively sets impedance parameters based on obstacle type and number, enabling soft compliance with alive obstacles and rigid compliance with inanimate obstacles. Impedance parameters with subscript \(o\) denote links between drones and obstacles, while subscript \(d\) denotes links between drones.}
\vspace{-6mm} 
\label{fig:main_image}
\end{figure}
\section{Introduction}
Among the most critical challenges in Swarm Robotics is enabling safe and intelligent navigation in dynamic, cluttered, and unpredictable environments. The ability to safely navigate around obstacles, whether humans or dynamic non-living objects, while maintaining swarm coordination, remains an open problem. While impedance control methods offer effective and fast obstacle avoidance by shaping compliant interactions, they typically rely on manually tuned parameters and are limited in adapting to diverse obstacle types or dynamic semantic contexts.
In this paper, we introduce \textbf{ImpedanceGPT}, a novel system that combines a Vision-Language Model (VLM) with Retrieval-Augmented Generation (RAG) method for enhanced navigation of a swarm of mini-drones in dynamic environments. The system leverages a custom database of environmental scenarios, enabling RAG to retrieve scenario-specific impedance parameters in real time. By considering a top-down view of the environment, the system dynamically adapts to different obstacles, ensuring robust and safe navigation for drones.
The key contributions of this work are as follows:
\begin{itemize}
    
    \item A novel impedance control that accounts for both dynamic alive (human) and dynamic inanimate (non-living) obstacles.
    \item A novel integration of VLM with RAG to provide semantic understanding and contextual decision-making for dynamic swarm navigation.
    \item An enhancement to impedance control systems by providing real-time adaptability to changing environmental conditions and obstacle types.
    \item The development of a custom database that allows RAG to generate impedance parameters based on scenario-specific inputs or visual cues, improving adaptability to diverse environments.
     
\end{itemize}
This work addresses the existing challenge of adapting impedance control to changing and unpredictable environments, particularly with respect to the handling of both dynamic alive and dynamic inanimate obstacles. Our proposed system provides a significant step forward in swarm robotics, contributing to more intelligent, adaptive, and robust autonomous navigation in real-world scenarios.
\begin{figure*}[t!]
\centering
\includegraphics[width=0.8\linewidth]{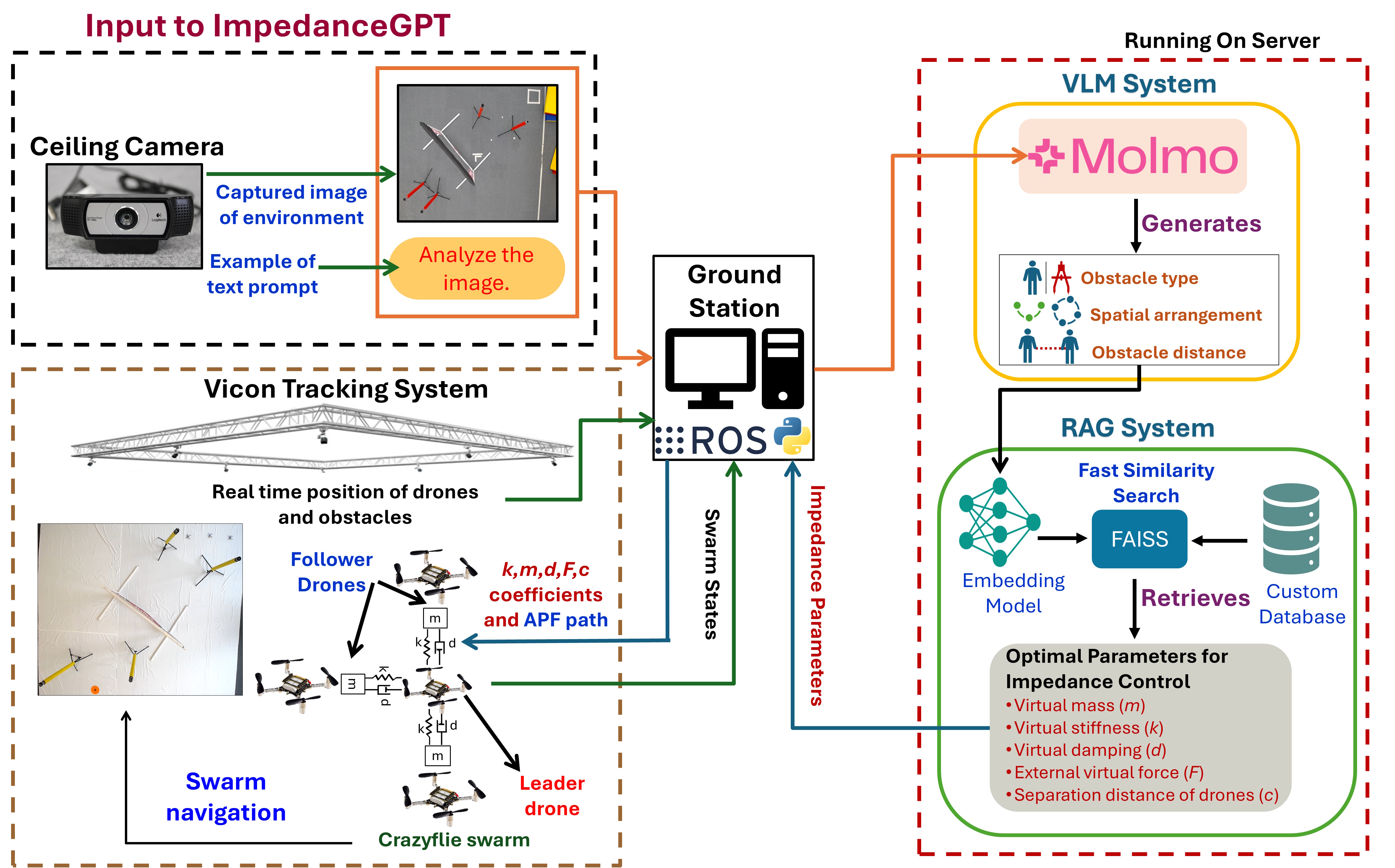}
\caption{System architecture of the ImpedanceGPT. The system transmits a top-down view from a ceiling- or drone-mounted camera, along with a user request, to Molmo. Molmo identifies obstacle types, distances, and arrangements, and then passes this information to the RAG. The RAG framework searches for and retrieves impedance parameters corresponding to the scenario that best matches Molmo’s description.}
\label{fig:architecture}
\vspace{-5mm}
\end{figure*}
\section{Related Works}
The field of safe and intelligent navigation in Swarm Robotics has significantly advanced; however, robust formation control in dynamic environments, while ensuring safe interaction within human-shared spaces, remains challenging \cite{Paper_01}. Active compliance methods, particularly impedance control, can effectively address these limitations by dynamically adjusting impedance parameters, primarily stiffness and damping, to facilitate safe and compliant behavior \cite{Paper_02}.
Impedance control for manipulators introduced by Hogan \cite{Paper_10}  is also widely recognized for enhancing path-following accuracy and managing aerial robotic manipulation tasks. Ahmed et al. \cite{Paper_03} introduced an Online Impedance Adaptive Control (OIAC) strategy, achieving superior UAV trajectory tracking compared to traditional PID and Model Reference Adaptive controllers. Similarly, Lippiello et al. \cite{Paper_04} presented a Cartesian impedance control framework that enables drones to compliantly interact with external disturbances and allows manipulators to handle contact forces during aerial manipulation tasks. Further studies \cite{Paper_05,Paper_06,Paper_07,Paper_08} developed impedance control frameworks specifically tailored for aerial manipulators, where impedance behavior is applied at the manipulator's end-effector, while the UAV itself remains rigid to maintain flight stability. Additionally, Rashad et al. \cite{Paper_09} explored Vision-Based Impedance Control (VBIC) for UAV contact inspection tasks. However, these studies primarily focus on compliant physical interaction, with many relying on manually preselected impedance parameters or lacking validation beyond simulated environments. Intelligent impedance control was introduced in
\cite{Paper_11}. Tsykunov et al. \cite{Paper_12} proposed the concept of virtual impedance links to achieve life-like and safe navigation of swarm of micro-quadrotors. Building on this foundation, Fedoseev et al. introduced tactile display DandelionTouch \cite{Paper_13}, demonstrating the potential for drone swarms to adapt the topology of virtual impedance links. Darush et al. \cite{Paper_14} further demonstrated improvements in swarm stability and safety using virtual impedance links. Khan et al. \cite{Paper_15} combined Artificial Potential Field (APF) algorithm with impedance control for navigation of drone swarm, although dynamic adjustment of impedance parameters relative to obstacle characteristics was not explored. Moreover, Zafar et al. \cite{Paper_16} emphasized the significance of task-specific, impedance-based navigation utilizing virtual links within heterogeneous swarm configurations. 
Integrating impedance control with models that provide spatial and semantic understanding of the environment could further enhance the navigation capabilities.
In this context, Vision-Language Models (VLMs), particularly those leveraging Vision-Based Transformers (VBTs), represent a pivotal method for augmenting visual perception and decision-making in robotic systems, significantly benefiting UAV navigation. For instance, Zhang et al. \cite{Paper_17} proposed VLM-Social-Nav, a socially aware robot navigation framework that employs VLMs to evaluate and select navigation trajectories. Similarly, Ye et al. \cite{Paper_18} combined VLMs with the Rapidly-exploring Random Tree (RRT) algorithm to plan paths that are not only physically feasible but also semantically meaningful. Model like Molmo-7B-O \cite{Paper_19}, capable of processing the multimodal data, has the potential to further advance autonomous decision-making in complex, dynamic environments.
Additionally, Sentence-BERT \cite{Paper_20} and similar models have been widely adopted to help robotic systems better understand environmental contexts by generating efficient text embeddings. Building upon SwarmPath \cite{Paper_15}, our current research integrates impedance control with Molmo, coupled with a RAG framework, to propose a unified system capable of predicting impedance parameters tailored to specific scenarios. This system, which can be integrated with various path planners, facilitates real-time semantic understanding for swarm navigation in complex environments, directly addressing existing research gaps where VLM-driven impedance control remains largely unexplored.


\vspace{-2mm} 
\section{ImpedanceGPT Technology}
\subsection{System Overview}
\vspace{-1mm} 
The \textbf{ImpedanceGPT} framework, illustrated in Fig.~\ref{fig:architecture}, consists of two primary components: the VLM-RAG system for impedance parameter estimation and a high-level control system integrating an Artificial Potential Field (APF) path
planner with impedance control. Inspired by the approach in \cite{Paper_15}, our implementation differs by employing a real drone as the leader. The VLM-RAG system processes a top-down view of the environment using the Molmo-7B-O BnB 4-bit model to analyze and extract key features, including the number, type (dynamic inanimate or dynamic alive), and spatial distribution of obstacles. The extracted
information is converted into a vector format and processed by the RAG system, which performs a similarity search within a custom database to retrieve optimal impedance control parameters. These parameters ensure a collision-free
path and include virtual mass ($m$), stiffness ($k$), damping ($d$), external virtual force ($F$), and separation distance ($c$) between the center of masses of the drones. Communication between the VLM-RAG framework running on the server and the drones, as well as real-world interactions, is managed via the ROS2 framework.

\vspace{-1mm}  
\subsection{Integrating VLM for Obstacle Identification and Spatial Analysis}
The VLM is employed to identify obstacles and determine their spatial arrangement. Visual inputs from a ceiling-mounted camera in the drone arena are processed to classify obstacles, count them, and determine  their precise positions. This information is used to construct a semantic representation of the environment.
To ensure efficient real-time processing, we utilize the Molmo-7B-O BnB 4-bit model as the VLM agent. The extracted obstacle data is then passed to the Retrieval-Augmented Generation (RAG) system, which retrieves suitable impedance control parameters based on the identified obstacle configuration. This enhances the swarm’s navigation and adaptability in both static and dynamic environments.

\subsection{Custom Database for Environmental Scenarios}
In order to build the RAG system, in this work a custom database is developed. This database contains optimal impedance parameters, including virtual mass ($m$), stiffness ($k$), damping ($d$), external virtual force ($F$), and separation distance ($c$) between the center of masses of the drones, for 40 different manually designed scenarios. Each scenario represents an indoor space with varying number, type, and spatial arrangement of obstacles.
The database is created by first simulating two and then four follower drones in a Gym PyBullet drone simulation environment, using APF as the global path planner. 

\subsubsection{\textbf{Handling Dynamic Alive and Dynamic Inanimate Obstacles}}
In this work, humans are considered as “soft” (alive) obstacles, while cylindrical obstacles are categorized as “hard” (inanimate) obstacles. The impedance behavior varies based on the obstacle type:
\begin{itemize}
    \item \textbf{Soft (alive) obstacles:} Low stiffness, moderate damping, and higher mass are required to ensure stable interactions and soft compliance with humans.
    \item \textbf{Hard (inanimate) obstacles:} High stiffness, high damping, and lower mass are needed to enable rapid response and rigid compliance.
\end{itemize}
Following these considerations, we recorded only the optimal impedance parameters from the PyBullet simulation that met the criteria: \textbf{large deflection and soft compliance} for soft obstacles and \textbf{small deflection and rigid compliance} for hard obstacles.
Table \ref{tab:impedance_ranges} shows the range of these parameters for different obstacles. The lower thresholds for these parameters were determined through real-world experiments. 
\renewcommand{\arraystretch}{1.3}
\newcolumntype{P}[1]{>{\centering\arraybackslash}p{#1}}


\begin{table}[]
\centering
\caption{\textsc{Optimal Parameters for Impedance Control}}
\begin{tabular}{|l|c|c|}
\hline
\multicolumn{1}{|l|}{\textbf{Parameter}}                                      & \textbf{\begin{tabular}[c]{@{}c@{}}Dynamic Inanimate \\ Obstacle\end{tabular}} & \textbf{\begin{tabular}[c]{@{}c@{}}Dynamic Alive \\ Obstacle\end{tabular}} \\ \hline
\textbf{Virtual mass (kg)}  & 1 - 1.5 & 3 - 7\\ \hline
\textbf{\begin{tabular}[c]{@{}l@{}}Virtual stiffness (N/m)\end{tabular}}    & 7 - 10 & 0.1 - 0.9 \\ \hline
\textbf{\begin{tabular}[c]{@{}l@{}}Virtual damping (Ns/m)\end{tabular}} & 3 - 5 & 1 - 2 \\ \hline
\textbf{\begin{tabular}[c]{@{}l@{}}External virtual force (N)\end{tabular}} & 0.4 - 0.7 & 0.2 - 0.45 \\ \hline
\textbf{\begin{tabular}[c]{@{}l@{}}Separation distance (m)\end{tabular}}    & 0.2 - 0.5  & 0.6 - 0.9 \\ \hline
\end{tabular}
\label{tab:impedance_ranges}
\vspace{-5mm}
\end{table}

\subsection{Retrieval-Augmented Generation (RAG) for Impedance Parameter Generation}
Retrieval-Augmented Generation (RAG) systems are typically used to enhance the response generation of Large Language Models (LLMs) by providing access to external knowledge bases. In this work, we implement a Naïve RAG approach, which utilizes a sentence transformer to create a text-embedded vector database and employs Facebook AI Similarity Search (FAISS) as the retrieval agent. 
The RAG system developed in this work processes the text output of the VLM model Molmo as a query, embeds it, and then uses FAISS to retrieve the optimal impedance parameters that best match the scenario described in the query. FAISS computes similarity between the query and the embedded vectors in the database using the Euclidean distance, given by the following expression:
\begin{equation}
d(X, Y) = \sqrt{\sum_{i=1}^{n} (X_i - Y_i)^2},
\end{equation}
where \( X = (X_1, X_2, ..., X_n) \) is the query embedding, \( Y = (Y_1, Y_2, ..., Y_n) \) is the stored embedding, and \( n = 384 \) is the embedding dimension.

\subsection{Impedance Control for Obstacle and Collision Avoidance in Drone Swarm Navigation}
Impedance control is integrated with APF algorithm to enhance swarm coordination and collision avoidance in drone navigation. Follower drones maintain connectivity with the leader drone through virtual impedance links modeled as a mass-spring-damper system, described by the equation:
\begin{equation}
    m\Delta\ddot{x} + d\Delta\dot{x} + k\Delta x = F_{\text{ext}}(t),
\end{equation}
where $\Delta x$, $\Delta \ddot{x}$, $\Delta \ddot{x}$  are the differences between the current and desired position, velocity, and acceleration, respectively, $m$ is the virtual mass, $d$ is the virtual damping, $k$ is the virtual stiffness, and $F_{\text{ext}}(t)$ is the virtual external force from the leader drone. Additionally, drones also dynamically form temporary virtual impedance links with nearby obstacles, governed by:
\vspace{-3mm}
\begin{equation}
    \Delta x_{\text{drone},n} = k_{impF} \cdot r_{imp},
\end{equation}
where $\Delta x_{\text{drone},n}$ is the difference between the obstacle and the $n_{th}$ drone; $n$ being the total number of drones, $r_{imp}$ defines the local deflection radius (0.65 m) and $k_{impF}$ adjusts the avoidance force based on drone velocity (0.45), enabling drones to temporarily diverge to avoid obstacles before rejoining trajectory of the leader drone.
\section{Experimental Setup}
In total, seven experiments were performed to evaluate the efficacy of the proposed ImpedanceGPT framework for the semantic understanding, obstacle identification, and intelligent navigation in both static and dynamic along with sparse and cluttered environments. These experiments also covered unseen dynamic scenarios, for which no pre-existing database was available, to further assess the framework’s robustness. The top-down images for each experiment were captured in daylight conditions to ensure consistent lighting and reliable image analysis. Multi-colored obstacles were placed within the environment to test the semantic classification capabilities of the VLM model and to check for any potential bias.

The VLM-RAG framework was executed on a high-performance PC equipped with an RTX 4090 graphics card (24GB VRAM) and an Intel Core i9-13900K processor to ensure smooth and efficient model inference. Due to memory constraints, the quantized Molmo-7B-O BnB 4-bit model was used.
For the real-world implementation of the system, Crazyflie 2.1 drones were utilized, allowing practical testing in dynamic environments. The VLM-RAG system was given the following text prompt for obstacle detection and information retrieval: \textit{“Analyze the drone arena image and identify all cylindrical stands or humans as obstacles in the scene. Count the total number of cylindrical stands or humans, specifying the number of obstacles before the gate and the number of obstacles after the gate. If there are no obstacles before the gate, mention that explicitly. Calculate the relative distances between each obstacle and describe their spacing as ‘closely spaced' or ‘widely spaced.' The spacing should be based on the distance between their feet in the image.”} The spacing between obstacles is determined based on the Euclidean distance computed from the normalized coordinates provided by Molmo. If the distance exceeds 20 grid cells, the obstacles are categorized as \textit{widely spaced}; otherwise, they are considered \textit{closely spaced}.
\vspace{-1mm}
\section{Experimental Results}
\subsection{Evaluation of VLM-RAG Framework}
The experimental setup represents an indoor environment, where lighting plays a crucial role in the performance of the VLM-RAG system. 
\begin{figure}[t!]
\centering
\includegraphics[width=1\linewidth]{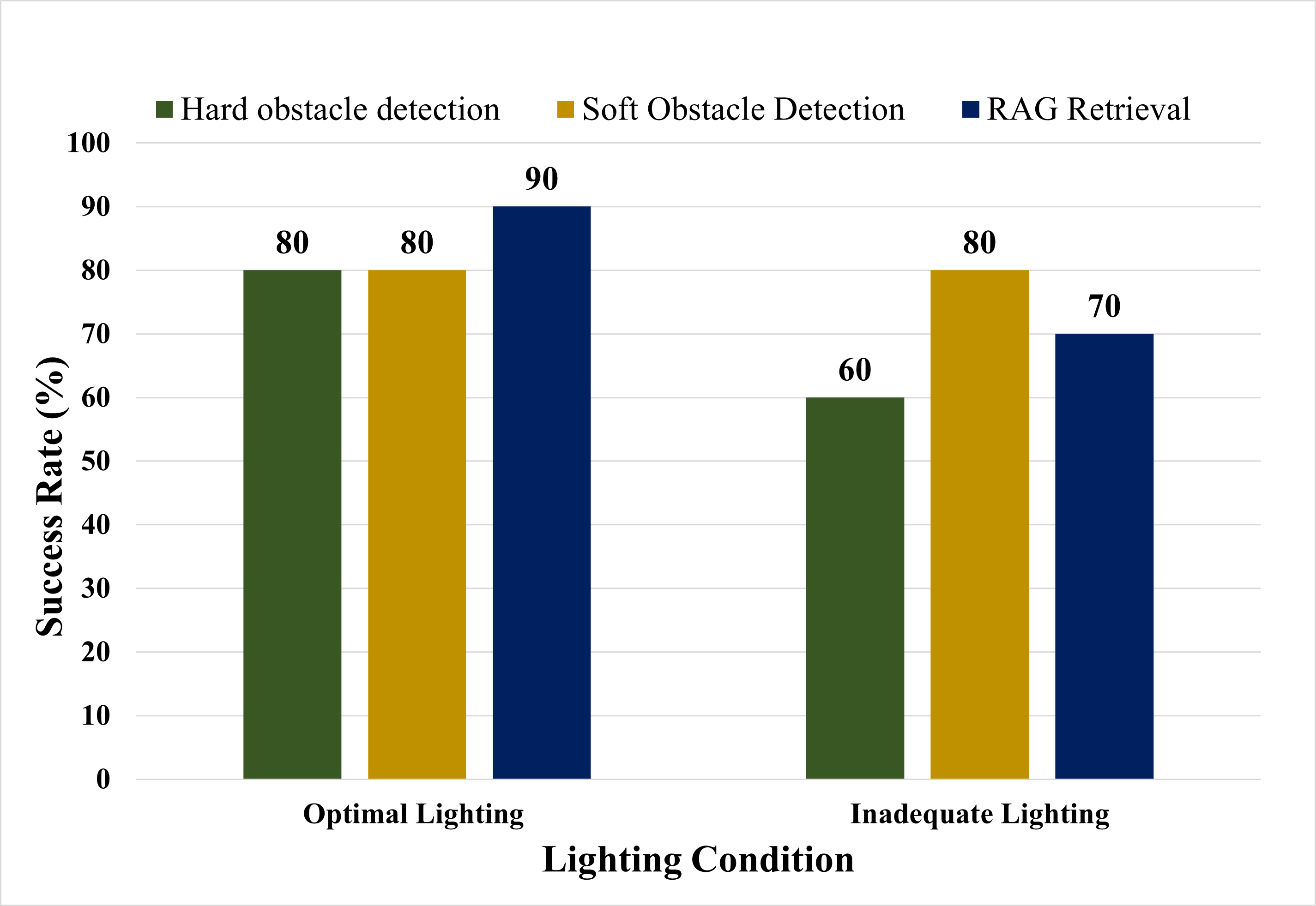}
\caption{Performance of VLM-RAG system under varying lighting conditions. \textit{Optimal lighting} denotes bright and well-illuminated environments, while \textit{inadequate lighting} corresponds to dim or poorly lit scenarios.}
\label{fig:VLM-RAG Performance}
\vspace{-8mm}
\end{figure}
A total of seven experiments were conducted, with each experiment repeated three times. The VLM-RAG system achieved an 80\% success rate in both the detection and retrieval of soft and hard obstacles under optimal lighting conditions. However, under inadequate lighting, the system struggled to identify obstacles of complex shape and close proximity, resulting in a reduced success rate of 60\%, as shown in Fig.~\ref{fig:VLM-RAG Performance}. The results are averaged across multiple scenarios in our experimental setup. While lighting conditions do affect performance, the model consistently detects obstacles and determines their positions regardless of color or placement. The computational time for each experiment ranged from 5 to 8 s.


\subsection{Results in Static Environment}

The experiments carried out in a static environment involved densely packed obstacles, creating challenging navigation conditions for the drones.

\begin{figure}[h]
\centering
\includegraphics[width=1\linewidth]{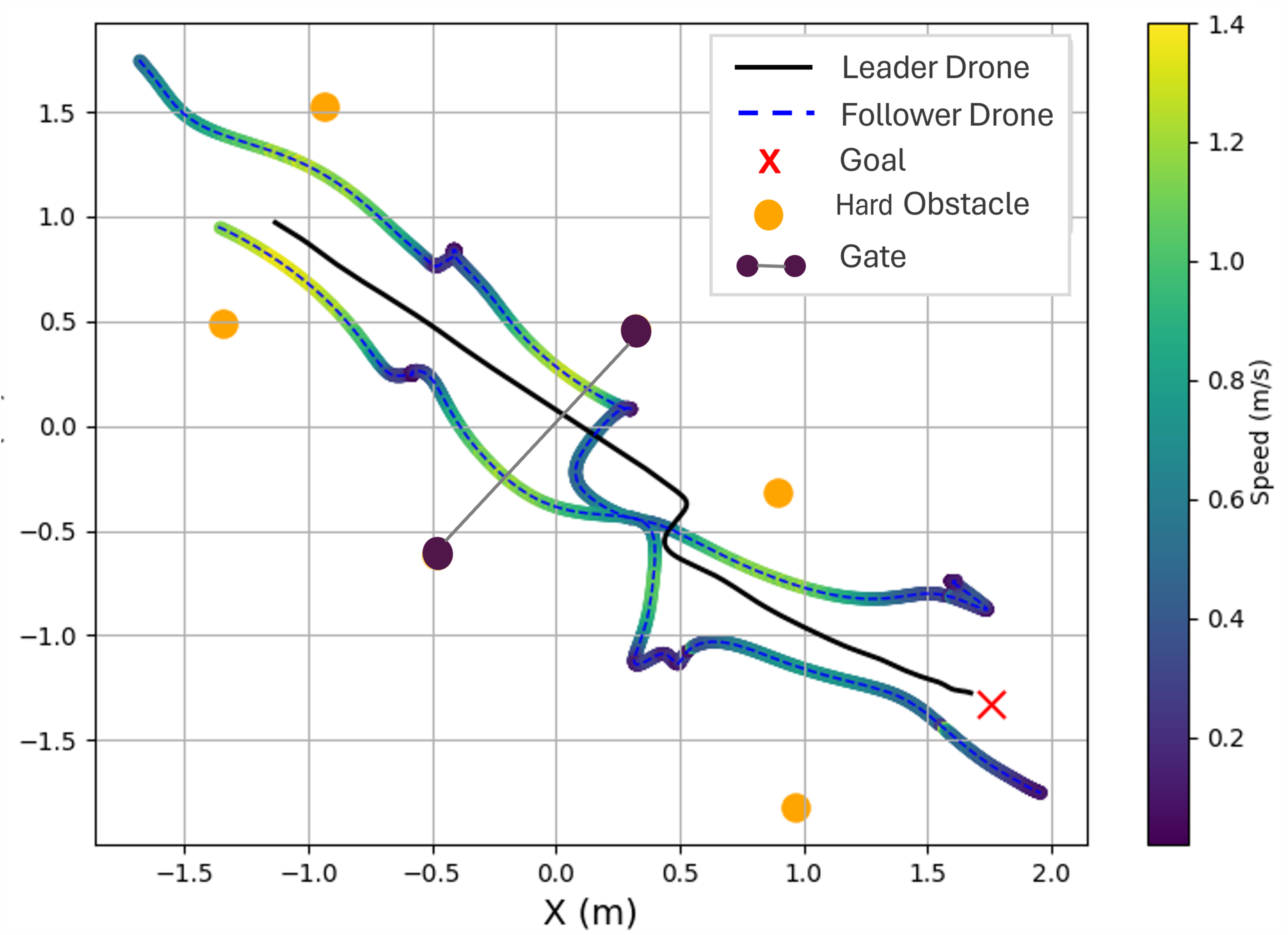} 
\caption{Scenario 1. Environment with 4 static hard obstacles and a rectangular gate.}
\label{fig:exp1}
\vspace{-4mm}
\end{figure}

\begin{figure}[h]
\centering
\includegraphics[width=1\linewidth]{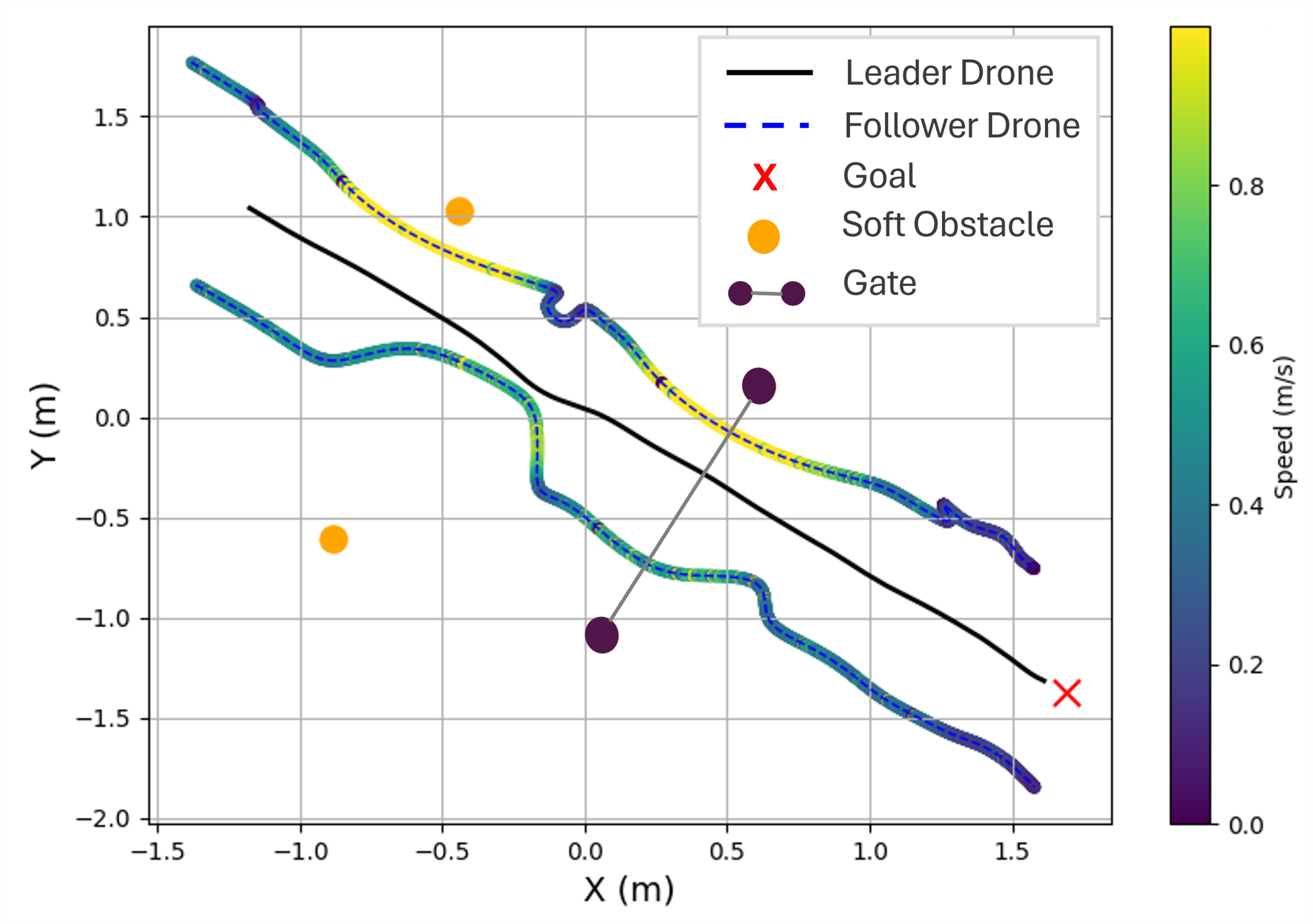} 
\caption{Scenario 2. Environment with 2 static soft obstacles and a rectangular gate.}
\label{fig:exp3}
\vspace{-2mm}
\end{figure}

Two representative scenarios are described below:
\begin{itemize}
    \item \textbf{Scenario 1, Static:} This scenario consists of four hard obstacles with a rectangular gate positioned within the environment. The drones are required to navigate through the space while avoiding collisions and reaching the designated goal position. As shown in Fig.~\ref{fig:exp1}, the drones successfully completed the task without any collisions. Overlapping trajectories of the follower drones do not indicate collisions but represent their positions at different time steps.
    
    \item \textbf{Scenario 2, Static:} In this scenario, two individuals (soft obstacles) are positioned before a rectangular gate in a way that they obstruct the path of swarm. Despite this, the drones successfully maneuver around the humans and proceed without any incident (see Fig.~\ref{fig:exp3}).
\end{itemize}

A key observation from these experiments is the effect of obstacle type on the drone behavior. In Scenario 1, where only hard obstacles are present, the trajectories overlap significantly, indicating higher movement speeds. Conversely, in Scenario 2, where soft obstacles (humans) are introduced, the drones maintain a greater separation distance, adjusting their paths based on the optimal parameters for impedance control generated by ImpedanceGPT. Additionally, the maximum velocity recorded in Scenario 1 is 1.4 m/s, whereas in Scenario 2, the velocity is reduced to 0.7 m/s, reinforcing the system’s ability to ensure safe navigation by modulating speed according to obstacle type.

\subsection{Results in Dynamic Environment}
Experiments in dynamic environments introduced additional complexities, including moving obstacles and changing goal positions, to assess the adaptability of the drone swarm. Two such dynamic scenarios are specified below:

\begin{figure}[h]
\centering
\includegraphics[width=1\linewidth]{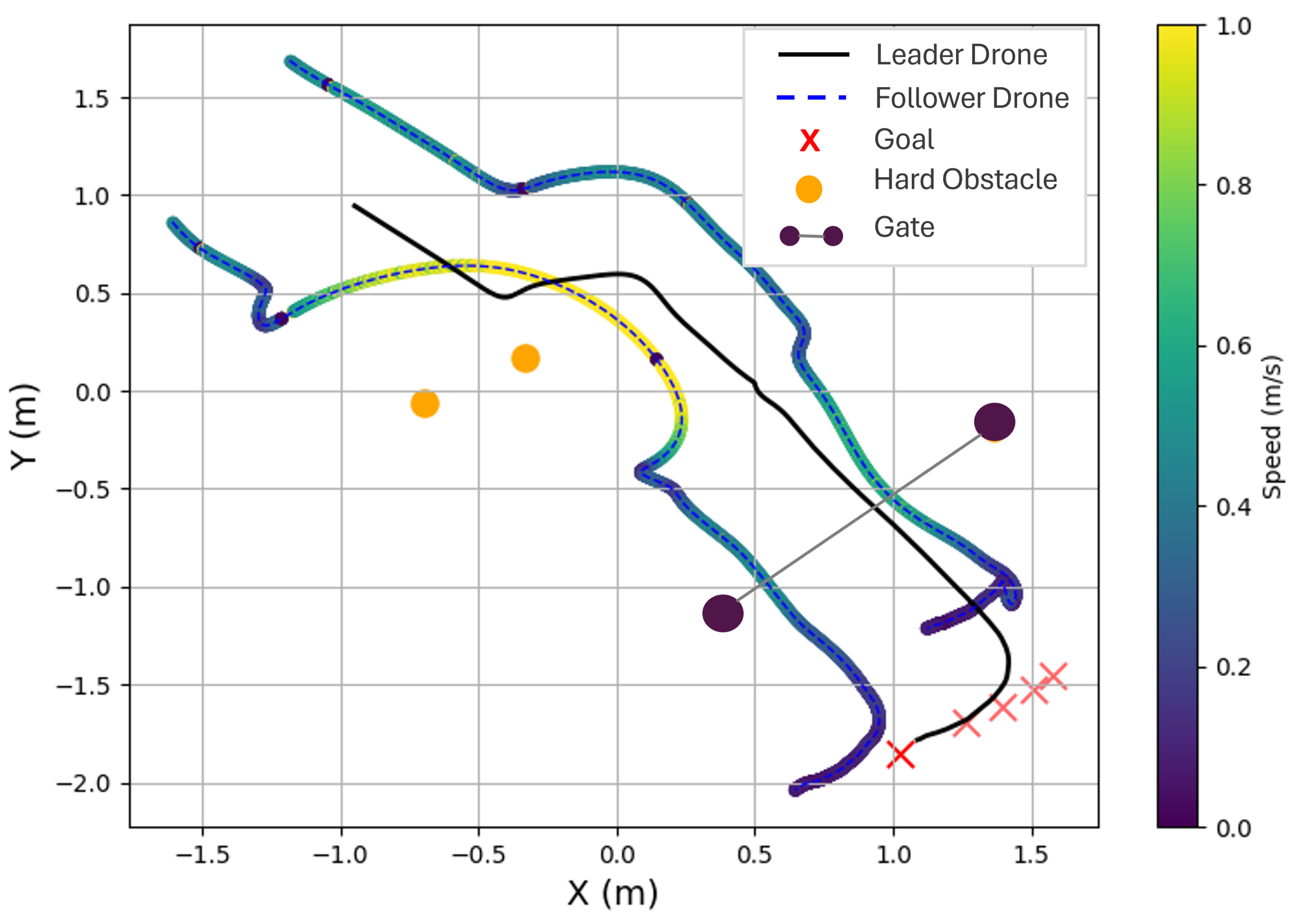} 
\caption{Scenario 3. Environment with 2 hard obstacles, a gate and moving goal position.}
\label{fig:exp2}
\vspace{-5mm}
\end{figure}

\begin{figure}[h]
\centering
\includegraphics[width=1\linewidth]{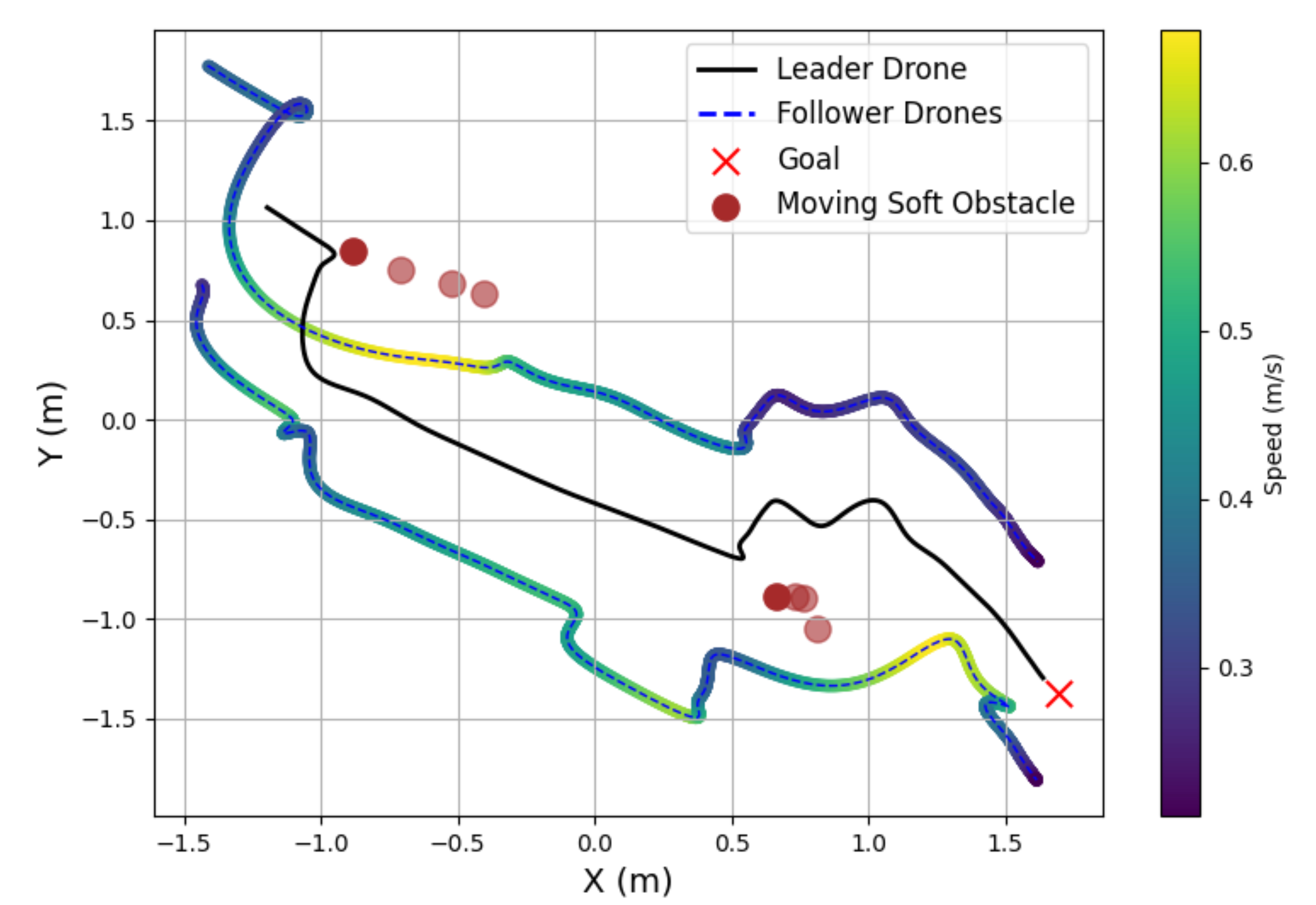} 
\caption{Scenario 4. Environment with 2 soft obstacles, one moving and one stationary.}
\label{fig:exp4}
\vspace{-5mm}
\end{figure}

\begin{itemize}
    \item \textbf{Scenario 3, Dynamic:} This scenario involves two follower drones guided by the leader drone, which actively tracks a moving goal position. The drones successfully avoid obstacles while continuously adjusting their paths in response to the environmental changes. As illustrated in Fig.~\ref{fig:exp2}, the leader drone and its followers dynamically navigate the environment, maintaining safe distances while ensuring smooth and adaptive motion towards the goal.
    
    \item \textbf{Scenario 4, Dynamic:} In this scenario, two humans serve as obstacles. One individual almost remains stationary in the path of the swarm, while the other moves freely. The swarm of drones successfully avoids the moving person and navigates past the stationary human, maintaining a safe distance and reducing speed as necessary to ensure collision-free navigation, as seen in Fig.~\ref{fig:exp4}.
\end{itemize}

A comparison of these dynamic scenarios further highlights the system’s ability to adapt motion based on the obstacle type. In Scenario 3, where drones navigate around a hard obstacle, they maintain a velocity of 1 m/s, enabling prompt movement through the environment. In contrast, in Scenario 4, where a human is dynamically moving, the velocity is further reduced to 0.6 m/s, demonstrating the system’s responsiveness in ensuring safe navigation by modulating speed when encountering soft dynamic obstacles. This adaptive behavior is crucial for maintaining safety and efficiency in real-world swarm navigation.

\subsection{Limitations}
The operational workspace is directly influenced by the field of view (FOV), where larger fields enable broader environments and more complex mission scenarios. In our case, the effective FOV of the available camera was limited to a $2 \times 2$ meter grid, which restricted the spatial scale of the experiments. Despite this constraint, we designed and executed scenarios of increasing complexity while ensuring safe and reliable UAV navigation within the available space.

\section{Conclusion and Future Work}
We have developed \textbf{ImpedanceGPT}, a VLM-RAG-driven impedance control designed for adaptive swarm navigation in dynamic environments. Through a series of experiments, the system demonstrated its capability in obstacle classification, impedance-based motion adaptation, and intelligent navigation. The VLM-RAG framework achieved a success rate of \textbf{80\%} in optimal lighting condition. In both static and dynamic environments, drones exhibited higher velocities for hard obstacles, i.e., \textbf{1.4 m/s} and \textbf{1.0 m/s}, respectively, while for soft obstacles, their velocities dropped down to \textbf{0.7 m/s} and \textbf{0.6 m/s}, respectively, thus ensuring safe and adaptive navigation.

Future work will focus on enhancing generalization by transfer learning, improving real-time adaptability via multi-objective optimization, incorporating edge computing for reduced latency, and advancing human-swarm interaction through natural language processing and gesture recognition. 

\section*{Acknowledgements} 
Research reported in this publication was financially supported by the RSF grant No. 24-41-02039.

\bibliographystyle{IEEEtran}
\balance
\bibliography{ref}

\end{document}